\PassOptionsToPackage{sort}{natbib}
\documentclass[unnumsec,webpdf,modern,large]{oup-authoring-template}
\usepackage{pdfpages}
\makeatletter
\def\ps@opening{\let\@oddhead\@empty\let\@evenhead\@empty\def\@oddfoot{\hfil\thepage\hfil}\let\@evenfoot\@oddfoot}
\def\ps@headings{\let\@oddhead\@empty\let\@evenhead\@empty\def\@oddfoot{\hfil\thepage\hfil}\let\@evenfoot\@oddfoot}
\def\societylogo{}
\makeatother

\begin{document}

\firstpage{1}

\title[Item-mean surrogates]{Item-Mean Surrogates: Why Richer Personas Fail to Deliver Individual Fidelity in LLMs as Human Surrogates}

\author[1,$\ast$]{Daehwan Ahn\ORCID{0000-0002-0886-6381}}
\author[2]{Chengfeng Mao\ORCID{0009-0002-6679-7764}}
\author[3,$\ast$]{Dokyun ``DK'' Lee\ORCID{0000-0002-3186-3349}}

\address[1]{\orgdiv{College of Family and Consumer Sciences}, \orgname{University of Georgia}, \orgaddress{\street{Athens}, \postcode{30602}, \state{Georgia}, \country{USA}}}
\address[2]{\orgname{MIT Sloan School of Management}, \orgaddress{\street{Cambridge}, \postcode{02142}, \state{Massachusetts}, \country{USA}}}
\address[3]{\orgdiv{Questrom School of Business \& Computing and Data Sciences}, \orgname{Boston University}, \orgaddress{\street{Boston}, \postcode{02215}, \state{Massachusetts}, \country{USA}}}

\corresp[$\ast$]{To whom correspondence may be addressed. Email: \href{mailto:daehwan@uga.edu}{daehwan@uga.edu} and \href{mailto:dokyun@bu.edu}{dokyun@bu.edu}}

\abstract{LLMs are increasingly used to simulate and study human behavior. Such studies assume richer personas improve predictions of specific people’s answers. We test this premise across four public datasets. First, we split each person's answer to an item into the item mean (the average human answer to that item) and the person's deviation from the item mean. Much of the accuracy that surrogate studies report comes from predicting item means. LLM predictions explain little of the deviations. We call this pattern item-mean surrogacy. Second, we split each deviation into stable signal and noise using a dataset in which the same participants answered the same questions twice. Stable signal is the part of a deviation that repeated answers share. Noise is the part that changes. In this dataset, more than half of the variance in deviations persists across occasions. Most of what the LLM misses is stable signal rather than noise. Third, we split stable signal into the person signal, a person’s average deviation across questions, and the person-by-item signal, their remaining question-specific deviations. The person-by-item signal accounts for about 8.5 times as much variance as the person signal. The LLM recovers about a tenth of it. Richer personas, different models, and fine-tuning on human answers still leave most of the person-by-item signal unrecovered. Both fine-tuned models recover less of this signal than their untuned versions. These findings call for closer empirical scrutiny of LLMs as human surrogates.}


\maketitle

\section{Introduction}

Large language models (LLMs) are increasingly used to study and simulate human behavior in marketing, political science, psychology, economics, and other social sciences. Proposed uses include filling in a person's missing responses across tasks or surveys and simulating responses from niche or hard-to-reach populations. Claimed benefits are lower costs, faster iteration without recruitment, and synthetic respondents on demand. Argyle et al.\ \cite{argyle2023out} reported that an LLM could reproduce aggregate survey distributions across demographic subgroups and introduced ``silicon samples'' for model-generated respondents. Aher et al.\ \cite{aher2023using} reported replicating several classic behavioral experiments at the group level. Studies that predict specific people's answers give the model a description of each person, or persona, assuming richer descriptions improve predictions. Park et al.\ \cite{park2024generative} built agents from self-reports by about 1,000 people, including two-hour interviews. They reported that the agents reproduced participants' General Social Survey (GSS) answers 85\% as accurately as participants reproduced their own answers two weeks later.

Such studies claim to simulate specific people \cite{park2024generative,toubia2025twin,chan2025digital}. Fidelity measures how closely a surrogate's predictions reproduce each person's own answers \cite{park2024generative,toubia2025twin}. A person's answer to an item can be decomposed into the item mean, the average human answer to that item, plus the person's deviation from the item mean. We evaluate fidelity at three levels. Item-mean fidelity requires the average prediction for each item to match the human item mean. Distributional fidelity requires the predictions for each item to spread around their mean as the human answers do, with the same shape of distribution. Individual fidelity also requires each person's predicted deviation to match that person's own deviation. Shuffling the predictions for an item across people keeps item-mean and distributional fidelity but loses individual fidelity, because each prediction now goes to the wrong person.

We reviewed 77 papers that propose, deploy, or test LLM surrogates (SI Appendix, Tables~S1--S1b). Of the 52 empirical papers that apply surrogate predictions, 35 evaluated item-mean fidelity and no higher level under our criteria, 11 evaluated distributional fidelity, and 6 compared predictions with each participant's answers across multiple questions. All six average an error or agreement measure over answers, such as mean absolute error (SI Appendix, Table~S1b). A prediction that is the same for every participant can also score high on such measures. Consider using the same item mean as the prediction for every participant. The prediction's error on each answer is the person's deviation from the item mean. The average error is thus small whenever deviations are small. A correlation computed across questions can also be high for the item-mean prediction \cite{netzer2026synthetic}. Item means differ across questions, so the item mean predicts which questions get higher answers and which get lower ones. A high score on these measures therefore does not show that a surrogate uses information about the person. None of the six studies reports comparison with this item-mean prediction as a baseline. Our first question is whether surrogates predict each person's deviation better than the item-mean baseline does.

To show whether a surrogate distinguishes people, a correlation should be computed separately for each item, so the correlation does not depend on the item mean. The few studies that report this correlation find it low. Peng et al.\ \cite{peng2025megastudy}  built surrogates from participants' earlier survey answers to predict their answers in 19 new studies, which we call the Megastudy. Across its 164 items, the average correlation was approximately 0.197. Wang et al.\ \cite{wang2026twins} found a median of 0.028, and Stojanov et al.\ \cite{stojanov2026participant} found means from 0.08 to 0.29, depending on the persona. These low correlations have two possible explanations, and they lead to different ways to improve surrogates. If the surrogate lacks information about the person, richer personas or better models could raise the correlation. If part of each deviation is random, no predictor can match that part. We call the random part noise, and the rest stable signal. Repeated answers from the same people to the same items can separate the two. Park et al.\ and Toubia et al.\ collected such answers but compared surrogates with them on the answers themselves, not on the deviations. Our second question is how much of each deviation is stable signal. Stable signal has two parts. The person signal is the stable part of a person's average deviation across items: some people answer most items higher than other people do. The person-by-item signal is the stable part of the deviation on a particular item beyond that average. For example, one person may like jazz but dislike cilantro compared to most people. A persona describes the person in general, through demographics, personality measures, or earlier survey records. It can inform the person signal, but it can inform the person-by-item signal only when its information relates to the item. This difference may explain why richer personas help little. Persona variables explain less than 10\% of the variance in human answers for many tasks \cite{hu2024quantifying}. In the Megastudy, personas with each participant's earlier survey records improved accuracy only slightly over empty personas. Researchers have also varied prompts, tested different models, and fine-tuned LLMs on human answers \cite{toubia2025twin,kolluri2025finetuning,peng2025megastudy}. Our third question is whether any of these changes predicts more of the person-by-item signal.

We answer the three questions with four public datasets (Table~\ref{tab:datasets}). In Twin-2K-500 \cite{toubia2025twin}, 2,058 participants answered more than 500 questions on demographics, personality, cognitive ability, and economic preferences over four waves. The fourth wave repeated 88 questions from earlier waves, so the same participants answered the same items twice. The Megastudy drew its participants from the same pool. SocSci210 \cite{kolluri2025finetuning} collects 210 social science experiments with their own participants. ANES \cite{argyle2023out} is a political survey. In each dataset we score the predictions that the source study released, and for SocSci210 we generate them with the released checkpoints. Methods describe the studies we retain and the predictions we evaluate in each dataset.

First, LLM item means correlate with human item means, but LLM predictions match each person's deviation weakly (Results 1). Across the four datasets, the squared correlation between predicted and observed deviations is 1.8\% to 10.8\%. The item-mean baseline scores higher than the LLM in all four datasets. We call this pattern item-mean surrogacy, and such models item-mean surrogates. Second, the repeated answers in Twin-2K-500 show that stable signal accounts for 61.2\% of the variance in deviations from item means (Results 2). Most of the deviation that the LLM fails to predict is therefore stable signal rather than noise. Most of the stable signal is person-by-item signal. The LLM predictions recover about a tenth of the person-by-item signal, and participants' earlier answers to the same items recover 58\%. Third, in the Megastudy and Twin-2K-500, we varied the persona, prompt, sampling setting, and model, and compared two models before and after fine-tuning on human answers (Results 3). None of these changes recovers most of the person-by-item signal. Both fine-tuned models recover less of it than the same models without fine-tuning.

\begin{table*}[t!]
\caption{Datasets and their roles in the analysis.}
\label{tab:datasets}
\tabcolsep=0pt
\begin{tabular*}{\textwidth}{@{\extracolsep{\fill}}>{\raggedright\arraybackslash}p{0.15\textwidth}>{\raggedright\arraybackslash}p{0.18\textwidth}rr>{\raggedright\arraybackslash}p{0.16\textwidth}>{\raggedright\arraybackslash}p{0.16\textwidth}@{\extracolsep{\fill}}}
\toprule
Dataset & Design & Participants & Items or outcomes & Primary analysis & Role \\
\midrule
Megastudy \cite{peng2025megastudy} & Within-study crossed, 18 studies & 1,784 & 160 & 133 items, 1,728 participants & Persona and model comparisons \\
Twin-2K-500 Survey \cite{toubia2025twin} & Crossed, condition blocks, two occasions & 2,058 & 126 & 49 items, 2,057 participants & Human retest and second fine-tuned setting \\
SocSci210 \cite{kolluri2025finetuning} & Nested, 210 studies & $>$400K & 5,998 & 556 held-out outcomes, 55,751 participants & Additional test across studies \\
ANES \cite{argyle2023out} & Within-study crossed, 2016 election & 4,270 & 11 & 4 items, 4,256 participants & Additional test in political surveys \\
\botrule
\end{tabular*}
\begin{tablenotes}%
\item Datasets are hereafter referred to as the Megastudy, Twin-2K-500, SocSci210, and ANES. Participants and items or outcomes describe each full dataset. Primary analysis gives the items used for the de-meaned $R^2$ and the variance decomposition. These items have ordered, bounded scales with documented endpoints. Their answers are placed on a common 0--100 scale (Percent of Maximum Possible). In Twin-2K-500, 21 of the 49 primary items come from between-subject experiments in which each participant was assigned to one condition, so each participant answered 37 of the 49 items. Design terms are defined in SI Appendix, Table~S3a. Megastudy counts cover the 18 retained studies.
\end{tablenotes}
\end{table*}

\section{Results}

\subsection{1. LLM Predictions Correlate with Item Means but Lack Distributional and Individual Fidelity}

\begin{figure*}[t!]
\centering
\includegraphics[width=\textwidth]{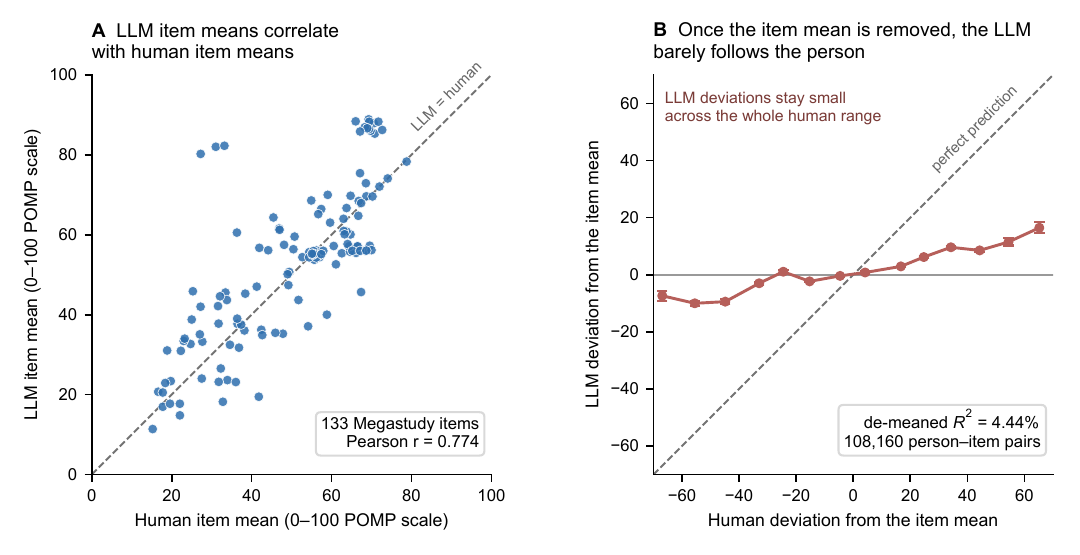}
\caption{LLM predictions correlate with item means but explain little of each person's deviation from them. (\textit{A}) LLM item means against human item means for the 133 Megastudy items on the Percent of Maximum Possible (POMP) scale, from full-persona GPT-4.1 at temperature zero (Pearson $r = .774$). The dashed line marks equal means. (\textit{B}) Each item's human mean is subtracted from both human answers and LLM predictions. Person--item pairs are grouped into bins by the human deviation. Each point is the mean LLM deviation in a bin, with a 95\% interval. The LLM deviations stay small across the range of human deviations. The dashed line marks perfect prediction. De-meaned $R^2 = 4.44\%$ ($n = 108{,}160$ person--item pairs).}
\figalttext
  [Scatter plots of LLM against human item means, and of binned LLM deviations against human deviations from the item mean.]
  {Panel A is a scatter plot of LLM item mean against human item mean, both on a 0--100 POMP scale, with one point for each of 133 Megastudy items and a dashed line marking equal means. Points lie mostly near the dashed line, with Pearson $r = .774$. Panel B plots LLM deviation from the item mean against human deviation from the item mean, each from about --60 to 60. Points are bin means with 95\% intervals, joined by a line, and a dashed diagonal marks perfect prediction. The binned LLM deviations stay small across the human range. De-meaned $R^2 = 4.44\%$ across 108,160 person--item pairs.}
\label{fig:itemmeans}
\end{figure*}

We place answers and predictions on a common 0--100 scale (Percent of Maximum Possible, POMP; Methods) to compare items with different response scales. The LLM predictions correlate with item means. Across the 133 Megastudy items, higher average human answers correspond to higher average LLM predictions (Pearson $r = .774$; Fig.~\ref{fig:itemmeans}\textit{A}). The LLM predictions lack distributional fidelity in three of the four datasets. The predictions for an item spread less widely than the human answers, as prior studies report \cite{gao2025take,bisbee2024synthetic,park2024diminished}. Across items, the median ratio of the LLM to the human standard deviation is 0.63 in the Megastudy, 0.58 in Twin-2K-500, 0.49 in SocSci210, and 0.99 in ANES (SI Appendix, Table~S6). The distribution of the predictions for an item also differs in shape from the distribution of the human answers, beyond this narrowing. In the Megastudy, the skewness of the predictions for each item correlates .397 with the skewness of the human answers. For comparison, we shift and narrow the human answers for each item to the mean and spread of the predictions, then round and clip them to the item's scale. The skewness of these transformed answers correlates .754 with the skewness of the human answers (SI Appendix, Table~S6b).

Individual fidelity requires the LLM to predict each person's deviation from the item mean. To test this, we subtract each item's average human answer from both the human answers and the LLM predictions (Fig.~\ref{fig:splits}, step 1). We call the squared Pearson correlation between these deviations across person--item pairs de-meaned $R^2$. It ranges from 0 to 1. Exact predictions score 1, but perfectly correlated deviations can also score 1 without matching in value. In the Megastudy, full-persona GPT-4.1 at temperature zero achieves 4.44\%, the highest de-meaned $R^2$ among the released models and settings (Fig.~\ref{fig:itemmeans}\textit{B}). The corresponding values are 5.28\% in Twin-2K-500, 10.77\% in ANES, and 1.83\% in SocSci210 (Table~\ref{tab:replication}, Panel~B).

This small association is stronger when each prediction is scored against the participant whose persona the LLM received. When we reassign each Megastudy participant's predictions to another participant in the same study, de-meaned $R^2$ falls from 4.44\% to at most 0.036\% across 10,000 reassignments (Methods).

The LLM predictions thus correlate with item means but lack individual fidelity. The LLM predicts from the person's persona and from its training data. The item-mean baseline predicts each answer with the average answer of the other participants to the same item (Methods), and does not use the person's persona. The baseline scores higher than the LLM in all four datasets by 1.1 to 2.9 points on 0--100 accuracy scales (Table~\ref{tab:replication}, Panel~A). ANES has the highest de-meaned $R^2$ of the four datasets, and the baseline scores higher than the LLM there too. These comparisons include Argyle et al.'s LLM predictions (ANES) and an LLM fine-tuned on 170 studies and evaluated on 40 studies held out from training (SocSci210). We use the source study's accuracy measure in three datasets and the Megastudy's normalized accuracy in ANES (Methods). Normalized accuracy is 100 minus the absolute difference between the prediction and the answer on the POMP scale. These models are therefore item-mean surrogates.

\subsection{2. Personas Miss How a Person Answers Particular Questions}

\begin{figure*}[t!]
\centering
\includegraphics[width=\textwidth]{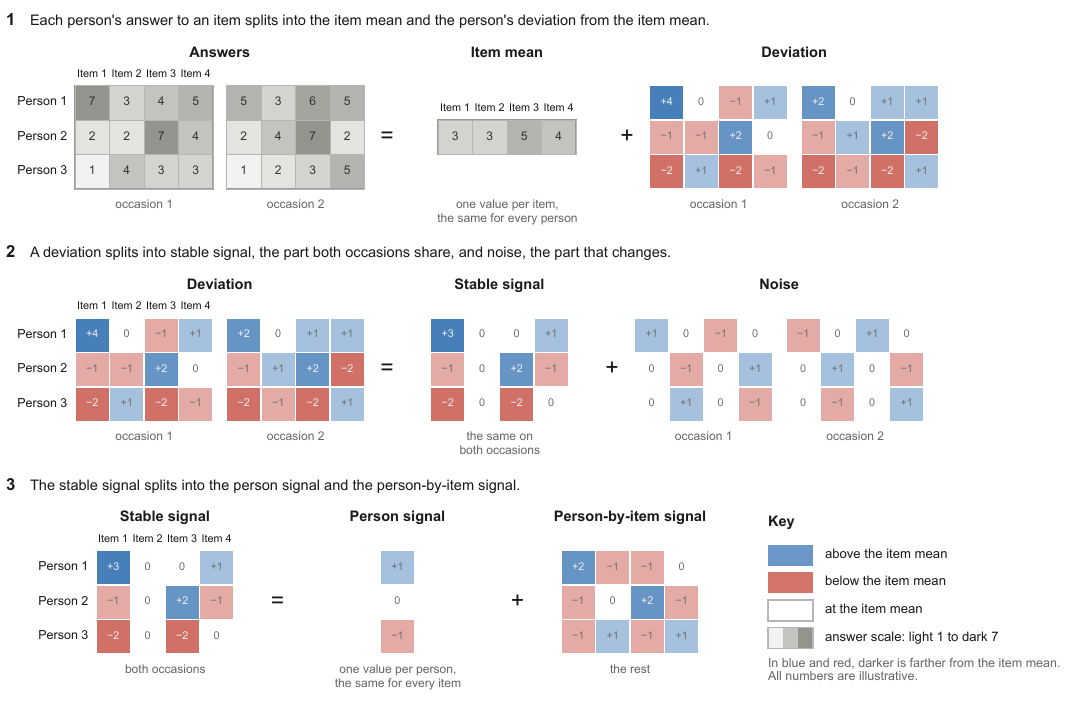}
\caption{Three splits of an answer, shown with illustrative numbers. Rows are people and columns are items. A part that repeats is drawn once. (1) Each person's answer to an item is the item mean plus the person's deviation from the item mean. The item mean is one value per item, the same for every person. (2) The deviation is stable signal plus noise. The stable signal is the part that both occasions share, so it is drawn once. Noise is the part that changes between occasions. (3) The stable signal is the person signal plus the person-by-item signal. The person signal is one value per person, the same for every item, and the person-by-item signal is the rest. Color marks a deviation and its parts: blue above the item mean, red below, and white at the item mean. Answers and item means are values on the answer scale, shown in gray that is darker for larger values. The numbers show only the arithmetic. We estimate the share of each part with a G-theory model fit by REML (Methods); Fig.~\ref{fig:variance}\textit{A} gives the shares in Twin-2K-500.}
\figalttext
  [Three splits of an answer, shown on grids of illustrative numbers with people as rows and items as columns.]
  {Three equations drawn with small grids of three people (rows, Person 1 to 3) and four items (columns, Item 1 to 4). Row 1: the answers on two occasions equal the item mean, drawn as a single row of four values, plus the deviations on the two occasions. Row 2: the deviations on the two occasions equal the stable signal, drawn as one grid for both occasions, plus the noise on each occasion. Row 3: the stable signal equals the person signal, drawn as a single column of three values, plus the person-by-item signal. Answers and item means are gray, darker for larger values on the answer scale; deviations and their parts are blue when positive, red when negative, and white when zero. A key at the lower right explains the colors.}
\label{fig:splits}
\end{figure*}

LLM predictions explain little of the deviations from item means. We ask how much of the deviation they miss is a stable signal and how much is noise. A single answer cannot separate stable signal from noise, but repeated answers across participants and items let us estimate both (Fig.~\ref{fig:splits}, step 2). Twin-2K-500 asked participants from the Megastudy's participant pool the same items twice, a median of about four weeks apart. We fit a generalizability-theory (G-theory) model by restricted maximum likelihood (REML) to these repeated answers (Methods). Stable signal accounts for 61.2\% of the variance in deviations from item means and noise for 38.8\%. Assuming noise is unrelated to predictions, no predictor can explain the noise. The share of stable signal is then the highest possible de-meaned $R^2$, our test--retest ceiling. The Twin-2K-500 surrogate's 5.28\% is about 9\% of that ceiling. Most variation it fails to predict is therefore stable signal rather than noise.

\begin{figure*}[t!]
\centering
\includegraphics[width=\textwidth]{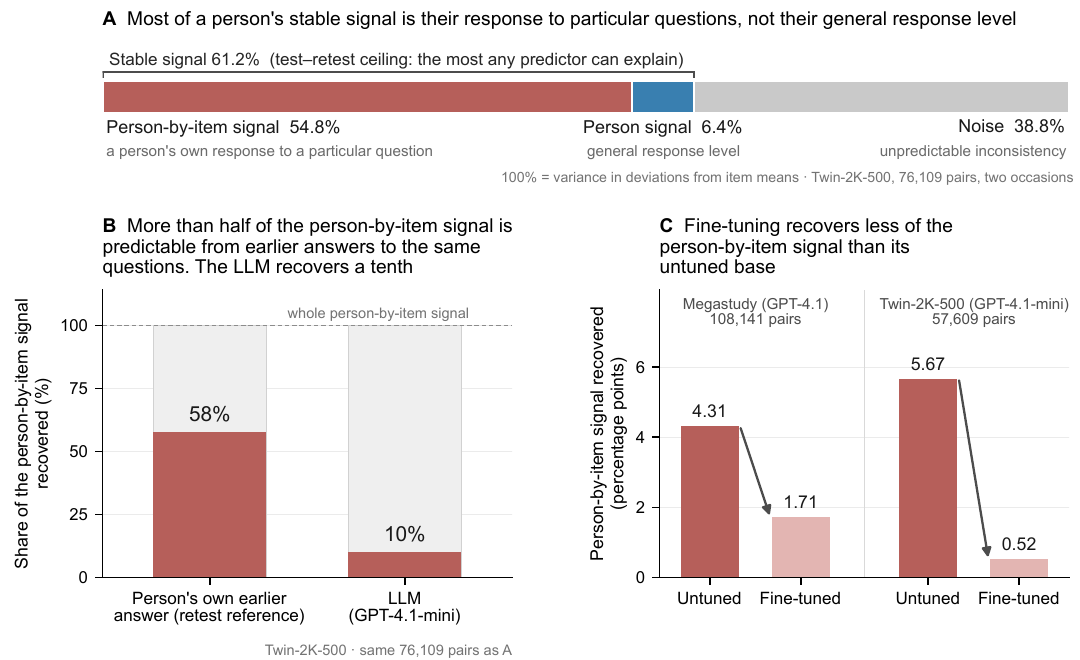}
\caption{Most of a person's stable signal is question-specific. More than half of the person-by-item signal is predictable from earlier answers to the same questions, which were unavailable to the LLM. The LLM recovers a tenth of it, and the fine-tuned models less still. (\textit{A}) Shares of the variance in deviations from item means in Twin-2K-500 (49 items, 2,057 participants, two occasions): person-by-item signal 54.8\%, person signal 6.4\%, and noise 38.8\% (1.3\% from a shift in the person's average deviation between occasions and 37.5\% from other changes between occasions; SI Appendix, Table~S2b). The two signals together, 61.2\%, are the test--retest ceiling. (\textit{B}) Share of the person-by-item signal recovered by the person's own earlier answer and by the LLM (GPT-4.1-mini, text format) on the same 76,109 Twin-2K-500 pairs: the earlier answer 58\% (95\% CI 57--59), from a recovery of 32.34 percentage points of the variance in deviations from item means, and the LLM 10\% (9--11), from 5.69 points (Methods; SI Appendix, Table~S2c). (\textit{C}) Recovery of the person-by-item signal by the untuned and fine-tuned models within each dataset, in percentage points of the variance in deviations from item means, on the pairs for which both models have predictions (Megastudy, 108,141 pairs; Twin-2K-500, 57,609 pairs, the fine-tuned model's coverage). The fine-tuned model recovers less than its untuned base by 2.60 points (95\% CI 2.18--2.96) in the Megastudy and by 5.16 points (4.58--5.85) in Twin-2K-500 (SI Appendix, Table~S2c).}
\figalttext
  [Variance shares in Twin-2K-500, the person-by-item signal recovered by earlier answers and the LLM, and fine-tuned recovery.]
  {Panel A is one horizontal bar showing shares of the variance in deviations from item means in Twin-2K-500: person-by-item signal 54.8\%, person signal 6.4\%, and noise 38.8\%. The two signals together, 61.2\%, are labeled the test--retest ceiling. Panel B shows two bars on a 0 to 100\% scale of the share of the person-by-item signal recovered, with a dashed line at 100\%: the person's own earlier answer recovers 58\% and the LLM (GPT-4.1-mini) recovers 10\%. Panel C shows paired bars of person-by-item signal recovered, in percentage points, for untuned and fine-tuned models: Megastudy 4.31 and 1.71, Twin-2K-500 5.67 and 0.52, with an arrow from each untuned bar to its fine-tuned bar.}
\label{fig:variance}
\end{figure*}

Step 3 of Fig.~\ref{fig:splits} splits the stable signal into the person signal and the person-by-item signal. Of Twin-2K-500's 61.2 percentage points of stable signal, 54.8 are person-by-item signal and 6.4 are person signal (Fig.~\ref{fig:variance}\textit{A}). Person-by-item signal is about 8.5 times as large as the person signal and accounts for most stable signal.

Participants' earlier answers to the same Twin-2K-500 items, unavailable to the LLM, provide a retest reference. We score earlier answers and LLM predictions against each person's later responses, using a separate calculation from de-meaned $R^2$ (Methods). Earlier answers recover 58\% of stable person-by-item signal and the LLM approximately 10\% (Fig.~\ref{fig:variance}\textit{B}). The two shares have different bases. The 61.2\% ceiling is the stable share of all variation in deviations from item means, and noise in the later answers is why it is below 100\%. The 58\% is the share of the stable person-by-item signal that an earlier answer recovers, and noise in the earlier answers is why it, too, is below 100\% (SI Appendix). More than half of this signal is therefore predictable when earlier answers to the same questions are available.

Megastudy participants answered each item once, so the Megastudy's share of stable person-by-item signal cannot be estimated directly. The person share is the part of the variance in deviations from item means that comes from each person's average deviation. In the Megastudy, it is estimated at 13.4\%. We assume that 61.2\% of this variance is stable, as in Twin-2K-500. We treat the person share as stable and subtract its 13.4 percentage points from 61.2. This leaves an estimated 47.8 percentage points of stable person-by-item signal (Methods). The full persona recovers 9.0\% of this estimated signal (SI Appendix, Table~S2c).

The remaining question is why personas recover only about a tenth of the person-by-item signal. Personas contain information recorded before the target question, such as demographics, personality measures, cognitive tasks, and economic preferences. These personas exclude the person's answer to the target question. In Twin-2K-500, the LLM recovers about a third of the person signal but a tenth of the person-by-item signal (SI Appendix). Recovery also varies across questions. On the ten items that ask about participants' own support for policies, the person-by-item signal recovered by the LLM accounts for 13.4 percentage points of the variance in deviations from item means, but only 0.4 points on the 29 items outside the policy task (SI Appendix, Table~S5c). The person-by-item signal is similar in these sets: 55.7\% and 49.7\% of the variance in deviations from item means within each set, respectively (SI Appendix, Table~S5d). Recovery therefore differs sharply across question sets despite substantial stable signal in both. This comparison does not establish whether personas lack relevant information or the model fails to use it.

\begin{table*}[t!]
\caption{Prediction accuracy with and without information about the target person, and prediction of deviations from human item means.}
\label{tab:replication}
\tabcolsep=0pt
\begin{tabular*}{\textwidth}{@{\extracolsep{\fill}}>{\raggedright\arraybackslash}p{0.26\textwidth}>{\raggedright\arraybackslash}p{0.40\textwidth}rr@{\extracolsep{\fill}}}
\toprule
\multicolumn{4}{@{}l}{\textit{Panel A. Prediction accuracy with and without information about the target person}} \\
\noalign{\vspace{2pt}}
Dataset and model & Accuracy evaluation & LLM & Item-mean baseline \\
\midrule
Megastudy, GPT-4.1 full persona, temperature 0 & 133 ordered items; normalized accuracy averaged with equal item weights & 75.80 & 76.91 \\
Twin-2K-500, GPT-4.1-mini text & 17 tasks; reconstruction of the source study's accuracy evaluation (human retest, 81.63) & 72.65 & 75.56 \\
ANES, GPT-3 & four attitude items; normalized accuracy & 75.3 & 76.6 \\
SocSci210, Socrates-DPO & 40 held-out source studies; 434,348 common pairs & 71.36 & 73.30 \\
\noalign{\vspace{6pt}}
\multicolumn{4}{@{}l}{\textit{Panel B. Prediction of deviations from human item means}} \\
\noalign{\vspace{2pt}}
Dataset and model & Primary evaluation & \multicolumn{2}{r}{De-meaned $R^2$} \\
\midrule
Megastudy, GPT-4.1 full persona, temperature 0 & 133 items, 1,728 participants, 108,160 pairs & \multicolumn{2}{r}{4.44\%} \\
Twin-2K-500, GPT-4.1-mini text & 49 items, 2,057 participants, 76,109 pairs & \multicolumn{2}{r}{5.28\%} \\
ANES, GPT-3 & four ordered attitude items, 4,256 participants, 14,469 pairs & \multicolumn{2}{r}{10.77\%} \\
SocSci210, Socrates-DPO & 556 retained ordered held-out outcomes, 55,751 participants, 320,544 pairs & \multicolumn{2}{r}{1.83\%} \\
\botrule
\end{tabular*}
\begin{tablenotes}%
\item Panel A compares LLM prediction accuracy with a predictor that uses other participants' answers to estimate each item mean, but no attributes of the target person. Accuracy is on a 0--100 scale on which 100 is exact agreement with the human answer. Accuracy definitions and aggregation differ across datasets, so the LLM and the baseline are compared within each row, not across rows. Panel B reports de-meaned $R^2$. It is the squared correlation between human answers and LLM predictions after the human item mean is subtracted from both. Full metric definitions, aggregation rules, and additional specifications are in the SI Appendix. Each row scores the surrogate that the source study built and released, in the specification used for its main analyses; for SocSci210, predictions we generated at temperature zero from the released checkpoint (Methods).
\end{tablenotes}
\end{table*}

\subsection{3. Richer Personas, Other Models, and Fine-Tuning Recover Little Person-by-Item Signal}

To measure how de-meaned $R^2$ changes with the specification, we vary the persona from empty to demographics to the full record of each participant's earlier surveys, and vary the prompt, the sampling settings, and the model. In the Megastudy, with the model held fixed at temperature zero, de-meaned $R^2$ rises from 0.10\% with an empty persona to 2.19\% with demographics and 4.44\% with the full persona. The full persona draws on each participant's earlier survey records, including psychological measures, cognitive tasks, and economic preferences. No other released specification exceeds 4.44\%, including chain-of-thought instructions, a higher temperature, and other models (GPT-5, DeepSeek, Gemini 2.5 Flash, and Gemini 3 Pro, 2.19\% to 3.44\%; SI Appendix, Table~S5). In Twin-2K-500, the other untuned specifications change the prompt, the output format, the temperature, or the model. Their de-meaned $R^2$ ranges from 3.87\% to 6.21\%, against 5.28\% for the primary one, and the best of them predicts a tenth of the 61.2\% share of stable signal (SI Appendix, Table~S3b).

As a reference for how much of a person's deviation the records can predict when a model also learns from other participants' answers, we fit a ridge regression to the records from which the personas were built. For each item, the ridge regression predicts the deviation from the item mean from the records, trained on the other participants' answers to that item and evaluated on participants held out of training (SI Appendix). With those answers available for training, the ridge regression scores a de-meaned $R^2$ of 9.00\% in the Megastudy and 15.35\% in Twin-2K-500. The LLM, given the persona and no answers to the item, scores 4.44\% and 5.28\%. We report the ridge regression as a reference for the signal in the records when such answers are available, not as a competitor on the same information.

Fine-tuning changes the model's weights rather than the persona or the prompt. Two datasets have a released fine-tuned model and its untuned base: in the Megastudy, Peng et al.'s GPT-4.1 fine-tuned on participants' earlier Twin-2K-500 answers against full-persona GPT-4.1; in Twin-2K-500, Toubia et al.'s fine-tuned GPT-4.1-mini against the untuned model (Methods). We score fine-tuning at all three levels of fidelity: the item means, the spread of the predictions for each item, and recovery of the person-by-item signal. A gain at one level does not show a gain at another. In the Megastudy, the spread of predictions of each item rises from 63\% of the human spread to 93\% (median across items; SI Appendix, Table~S7). Yet both fine-tuned models recover less of the person-by-item signal than their untuned bases, falling from 4.31 to 1.71 points in the Megastudy and from 5.67 to 0.52 in Twin-2K-500 (Fig.~\ref{fig:variance}\textit{C}). Matching the spread of human answers is different from predicting specific people's answers.

Neither fine-tuned model matches item means more closely. On the pairs common to each comparison, the correlation between LLM and human item means falls from .77 to .59 in the Megastudy and from .70 to .29 in Twin-2K-500 (SI Appendix, Table~S7). The mean absolute difference between LLM and human item means rises as well, from 8.6 to 20.3 points in the Megastudy and from 11.5 to 11.8 in Twin-2K-500.

To tell whether fine-tuned models predict participants in new studies as well as in the studies used in their training, we compare the SocSci210 surrogates on the 170 studies used in their training and on the 40 studies held out (Methods). With supervised fine-tuning, de-meaned $R^2$ is 7.57\% on the studies used in training and 0.73\% on the held-out studies. With preference optimization, the variant scored in Results 1, de-meaned $R^2$ is 3.65\% on the training studies and 1.83\% on the held-out studies (SI Appendix). An untuned GPT-4o-mini, which was not fine-tuned on any of the studies, scores 1.50\% on the training studies and 1.77\% on the held-out studies. Both fine-tuned models score well above the untuned model on the studies used in training, and about the same or lower on the held-out studies.

\section{Discussion}

\subsection{What the Decomposition Adds}

A model that predicts the item mean for every person can have low error and high agreement with the answers. Only a score on the deviations, such as de-meaned $R^2$, shows whether a model distinguishes people. On this score, the models in all four datasets are item-mean surrogates. In Twin-2K-500, the person-by-item signal is most of the stable signal. Earlier answers to the same questions show that more than half of it is predictable. The LLM recovers a tenth of it.

A persona records information about the person, such as demographics and earlier survey records, but not the person's answer to the target item. It can therefore inform the person-by-item signal only on items related to what it records. The results in Twin-2K-500 fit this account (Results 2). The LLM recovers more of the person signal than of the person-by-item signal. Within the person-by-item signal, it recovers far more on the ten policy items than on the other items. This comparison cannot distinguish whether the personas lack the information or the model fails to use it.

Fine-tuning brings the spread of the predictions closer to the human spread in the Megastudy but lowers recovery of the person-by-item signal in both the Megastudy and Twin-2K-500 (Results 3). The decomposition makes the distinction clear: matching how much answers vary across people is not the same as recovering the person-by-item signal.

\subsection{Connecting Prior Findings}

Two studies of the same surrogates reach opposite conclusions, because they use different scores. Toubia et al.\ built surrogates from the Twin-2K-500 participants' records and compared their accuracy with retest accuracy, the accuracy with which participants reproduce their own earlier answers \cite{toubia2025twin}. By this score, the surrogates performed well. Peng et al.\ used the same surrogates to predict the same participants' answers in new studies and found that the surrogates told participants apart poorly \cite{peng2025megastudy}. Accuracy gives a high score to a prediction close to the item mean. In our reconstruction of Toubia et al.'s evaluation, the item-mean baseline reaches 92.6\% of retest accuracy, and the LLM reaches 89.0\%. Peng et al.\ instead computed a correlation for each item across the participants who answered it. A prediction of the item mean cannot raise this correlation, so only a surrogate that tells people apart scores high on it.

Item-mean surrogacy also connects findings reported separately: weak persona effects \cite{hu2024quantifying,tadaymorocho2026assessing}, compressed response distributions \cite{gao2025take,bisbee2024synthetic,park2024diminished}, low scores on behavior-simulation benchmarks \cite{hu2025simbench}, LLMs describing response distributions more accurately than they simulate them \cite{meister2025distributional}, and aggregate accuracy that does not extend to individuals \cite{petrov2024limited,li2025structural,xie2026statistical,wang2026twins,stojanov2026participant}. Two recent evaluations come close to the item-mean comparison. Moon et al.\ found that for purchase quantity no model scored higher than a baseline that simply matched the pooled human distribution, and concluded that mean-based evaluation alone can be misleading \cite{moon2026beyond}. Kinzinger and Hartmann found that an LLM with no persona answers many items correctly because the items have a modal answer that the model predicts \cite{kinzinger2026synthetic}. Neither compares the surrogate, person by person, with an item mean that ignores the person. We do so in all four datasets.

The closest designs to ours reach similar conclusions. Netzer and Sambandam separate the between-question component of answers from the within-question, between-person component \cite{netzer2026synthetic}. Our decomposition goes further. With repeated human answers, we split the within-question component into stable signal and noise, and the stable signal into the person signal and the person-by-item signal. We then measure how much of each signal the LLM predictions recover.  Jia et al.\ built personas from the same people's earlier answers and evaluated held-out items per person, as we do, and advise using digital personas most cautiously for individual-level prediction \cite{jia2026digital}. They excluded majority-answer baselines as oracle references, because such baselines require the held-out human answers. In our setting those answers exist, and the comparison shows that the item mean alone scores higher than the surrogates.

\subsection{Validation Depends on the Claim}

Whether a surrogate must predict a person's deviations from the item mean depends on the use a researcher intends \cite{crockett2025dead,hullman2026heuristic,lin2025six,sen2025methodological}. A study estimating a population mean needs item-mean fidelity, not a reconstruction of every person. A study of distributions of answers also needs distributional fidelity, because it must reproduce how common different answers are. The surrogates here do not reproduce human response distributions. In the Megastudy, Twin-2K-500, and SocSci210, their answers are narrower than the human answers and differently shaped (SI Appendix, \textit{Distributional Compression} and \textit{Shape Distortion}, Figs.~S1--S2 and~S4, Tables~S6--S6b). Neither item-mean nor distributional fidelity shows that a surrogate can replace a specific person. A study that predicts specific people's answers needs individual fidelity. The surrogates here also lack individual fidelity (Results 1).

For studies that need only item-mean or distributional fidelity, recent methods improve LLM predictions. Krsteski et al.\ \cite{krsteski2025valid} combine prompting or fine-tuning with human-data-based rectification to reduce bias in population estimates. Anamnesis \cite{yu2026anamnesis} uses narrative backstories to produce opinion distributions closer to the human distributions than standard persona-prompting baselines. These approaches improve estimates of population averages and response distributions.

For predicting specific people's answers, recent methods add the person's own earlier answers. PersonaTwin \cite{chen2025personatwin} combines demographic, behavioral, and psychometric information with updates from a person's prior conversational responses, reporting improvements in personalized text generation and downstream prediction. The comparison with the person's earlier answers shows why such history can help: the person's own answers to the same questions contain person-by-item signal that the LLM does not recover from the persona records. New methods can report how much of the person signal and of the person-by-item signal they recover, as we do here.

\subsection{Conclusion}

The LLM surrogates we examined are item-mean surrogates. Most of a person's stable signal is the person-by-item signal, their response to particular questions, and the LLM recovered about a tenth of that signal. Richer personas improve prediction without recovering most of the person-by-item signal, and the two fine-tuned models recover less of it than the same models without fine-tuning. Standing in for a specific person also requires answering questions the person was never asked. For such questions, the person's own earlier answer, which recovered more than half of the person-by-item signal in Twin-2K-500, does not exist. In SocSci210, the fine-tuned models scored well above an untuned model on the studies used in training, but not on held-out studies. The evidence we reviewed does not yet establish individual fidelity, and the surrogates we tested do not show it.

\section{Materials and Methods}

\noindent The SI Appendix gives the notation guide, the full procedures, and the additional specifications.

\noindent\textit{Notation.} $r$ denotes the Pearson correlation. POMP (Percent of Maximum Possible) rescales each item's answers and predictions to a common 0--100 range using the item's documented endpoints. Let $p$ index participants, $i$ index items, and $\Omega$ denote observed matched person--item pairs. On the POMP scale, $z_{pi}$ and $\hat z_{pi}$ denote the human answers and model predictions, and $\bar z_{\cdot i}$ denotes the human item mean. De-meaned answers and predictions are $d_{pi}=z_{pi}-\bar z_{\cdot i}$ and $\hat d_{pi}=\hat z_{pi}-\bar z_{\cdot i}$. De-meaned $R^2$ is $\mathrm{Corr}^2_{(p,i)\in\Omega}(d_{pi},\hat d_{pi})$, the squared Pearson correlation pooled across all prediction--answer pairs of a dataset. Centering both series on the human item mean follows profile-similarity methodology \cite{cronbach1953assessing, furr2008framework, wood2016correlates} and is equivalent to group-mean centering in multilevel analysis \cite{enders2007centering}. The SI Appendix reports a variant that centers the model predictions on their own item mean. For the primary Megastudy specification, centering the predictions on their own item mean gives 5.70\%, and centering on the human item mean gives 4.44\%.

\subsection{Datasets and Primary Specifications}
We analyze four datasets (Table~\ref{tab:datasets}). Each dataset is scored with the surrogate its source study built and released, not with a surrogate of our own. Where a source study released more than one specification, we score the one used for its main analyses. In the Megastudy, this is the full-persona GPT-4.1 prompt at temperature zero, which is also the strongest of the 23 released specifications on de-meaned $R^2$ (SI Appendix, Tables~S5 and~S8). In Twin-2K-500, it is the text-format GPT-4.1-mini specification, whose seven other released variants range from 3.87\% to 6.21\% (SI Appendix, Table S3b). In SocSci210, it is the Socrates checkpoint after preference optimization, which also scores higher on the held-out studies than the supervised fine-tuning checkpoint. We generated its predictions at temperature zero, as in the primary Megastudy specification, so that sampling noise does not lower de-meaned $R^2$; the source study reports inference at temperature 0.6 with nucleus sampling (SI Appendix). ANES has one set of released predictions. The Megastudy \cite{peng2025megastudy} contains 164 target items from 19 studies, and Twin-2K-500 contains 126 items. For SocSci210 \cite{kolluri2025finetuning}, we generated predictions with the released Socrates checkpoints (Fine-Tuned Models). For ANES, we use the GPT-3 predictions released by Argyle et al.\ \cite{argyle2023out} for four attitude items. In all four datasets, the primary analysis uses only items with ordered, bounded response scales, so that every answer can be placed on the common 0--100 scale (POMP). We excluded nominal categories, computed scores, objective-correctness measures, open numeric estimates, and unresolved special codes, because they have no order or no documented bounds. This leaves 133 Megastudy items and 49 Twin-2K-500 items. In the Megastudy, we also followed Peng et al.\ in excluding one third-party creativity rating and three variables that have no outcome measure. One study lost all its items, so 160 items from 18 studies remained before the selection above. The Twin-2K-500 personas were built from the participants' earlier answers, but not from their earlier answers to the 49 primary items. We use those earlier answers as a retest reference in Results 2. The LLM never saw them (SI Appendix). The SI Appendix reports broader item sets, original-scale comparisons, and how we handle the ANES ideology code. 

\subsection{Research Ethics}
No new human-subjects data were collected and no participant is identifiable. The University of Georgia Human Research Protection Program determined that the project is not research involving human subjects as defined by DHHS and FDA regulations (Not Human Subjects Research determination, IRB ID PROJECT00014809).

\subsection{Item-Mean Baseline}
For each dataset, the item-mean baseline predicts a participant's answer with the mean answer of other participants to the same item, estimated by five-fold cross-fitting over participants, so that the target participant's answer never enters the mean. The baseline is scored with the same accuracy definition and the same answers as the LLM in that dataset. Accuracy definitions follow the source studies where one exists and are listed with their aggregation rules in the SI Appendix. Argyle et al.\ did not score the four ANES attitude items answer by answer, so for ANES we apply the normalized accuracy that Peng et al.\ and Kolluri et al.\ use. The SI Appendix also reports a leave-one-out per-person comparison.

\subsection{Variance Decomposition}
The variance decomposition uses restricted maximum likelihood (REML) fits with \texttt{lme4} \cite{bates2015fitting}, following generalizability-theory conventions \cite{cronbach1972dependability,brennan2001generalizability}.

\textit{Human answers.} In Twin-2K-500, answers from both occasions are decomposed with occasion as a fixed effect and person, item, person-by-item, person-by-occasion, and item-by-occasion as random effects. The person signal is the part of a person's average deviation across items that the two occasions share. The person-by-item signal is the part of the person's deviation on each item, after removing that average, that the two occasions share. Shares are expressed relative to the single-occasion variance in deviations from item means, so that the person, person-by-item, person-by-occasion, and residual components sum to 100\%.

\textit{Recovery.} For a predictor $\hat z$, we fit a model with person and item random effects and a residual to the human answers, to the predictions, and to their sum on the same pairs. The covariance between the human and predictor components is $[\mathrm{Var}(z+\hat z) - \mathrm{Var}(z) - \mathrm{Var}(\hat z)]/2$, component by component. Recovery within a component is $100\max(\mathrm{cov},0)^2/[\mathrm{Var}(\hat z)\,V_z]$ for that component, where $V_z$ is the human variance in deviations from item means. Recovery is the human variance in that component that is linearly predictable from the predictor's component, in percentage points of $V_z$. The residual component of this single-occasion model contains the person-by-item signal together with noise specific to that occasion. An earlier answer or a persona-based prediction is formed without the target occasion's answers, so under the response model (SI Appendix) it is uncorrelated with that noise. Its recovery of the residual therefore measures only the person-by-item signal, and we report it as the recovery of the person-by-item signal in Results. The recovery of the person signal and the recovery of the person-by-item signal come from separate fits, so they do not add up to de-meaned $R^2$. To express recovery as a fraction of the person signal or the person-by-item signal, we convert the recovery and that signal to POMP-squared units before dividing.

\subsection{Retest Reliability and the Megastudy Estimate}
The retest correlation is the pooled correlation between a participant's two answers to the same item after centering each occasion on its own item mean, across the 49 primary Twin-2K-500 items ($r_{tt} = .612$; 95\% CI $[.602, .623]$ from 2,000 person-cluster bootstrap replicates with item means recalculated in each replicate; all 2,058 participants; median retest interval about four weeks). Under the response model in the SI Appendix, the retest correlation equals the share of stable signal in the variance in deviations from item means. The Megastudy items were answered once, so their person-by-item share is estimated from the Twin-2K-500 value: $r_{tt}$ times the sum of the person and residual shares of the Megastudy human answers, minus the person share. The person share is the part of the variance from each person's average deviation. The estimated Megastudy shares and their sensitivity to $r_{tt}$ are reported in the SI Appendix, and Results 2 uses them to express the Megastudy recovery as a fraction of the estimated person-by-item signal.

\subsection{Reassigning Predictions to Other Participants}
Within each Megastudy study, the existing predictions were reassigned to other participants 10,000 times, keeping each participant's human answers fixed, and de-meaned $R^2$ was recomputed under each reassignment (SI Appendix).

\subsection{Fine-Tuned Models}
Peng et al.\ fine-tuned GPT-4.1 with supervised learning on participants' Twin-2K-500 answers and evaluated it on the Megastudy outcomes, which were not part of the training data \cite{peng2025megastudy}. We compare its temperature-zero predictions with the untuned full-persona GPT-4.1 predictions on the 108,141 pairs common to both. Toubia et al.\ lightly fine-tuned GPT-4.1-mini on 500 labeled personas \cite{toubia2025twin}. The fine-tuned model's predictions cover 1,557 participants, and we compare them with the text-format GPT-4.1-mini predictions on the 57,609 common pairs. Socrates \cite{kolluri2025finetuning} is Qwen2.5-14B fine-tuned on 170 of the 210 SocSci210 studies, with supervised fine-tuning on the response log-likelihood (SFT) or direct preference optimization (DPO). The remaining 40 studies are held out at the study level (SI Appendix).

\subsection{Statistical Reporting}
Point estimates are reported throughout, with person-cluster bootstrap 95\% confidence intervals where indicated (the retest reliability and recovery of each signal; SI Appendix, Table~S2c).

\section{Acknowledgments}
We used Claude Opus 5.5 and Claude Fable 5.1 (Anthropic) and GPT-6 Sol and GPT-6 Astra (OpenAI) to assist with drafting, editing, and reviewing text, proofreading, and writing reproducibility code. The authors reviewed and verified all content and take full responsibility for it.

\section{Supplementary Material}
The SI Appendix follows the references.

\section{Competing interest}
The authors declare no competing interests. None of the authors have financial investments or start-ups related to LLM-based human digital twins.

\section{Funding}
The authors declare no funding.

\section{Author contributions}
D.A., C.M., and D.L. designed research; D.A., C.M., and D.L. performed research; D.A., C.M., and D.L. analyzed data; and D.A., C.M., and D.L. wrote the paper.

\section{Data availability}
Analysis code and derived data tables are deposited at the Open Science Framework (https://osf.io/42j69). The repository will be made public and assigned a DOI.

\bibliographystyle{unsrt}
\bibliography{references}

\clearpage
\includepdf[pages=-]{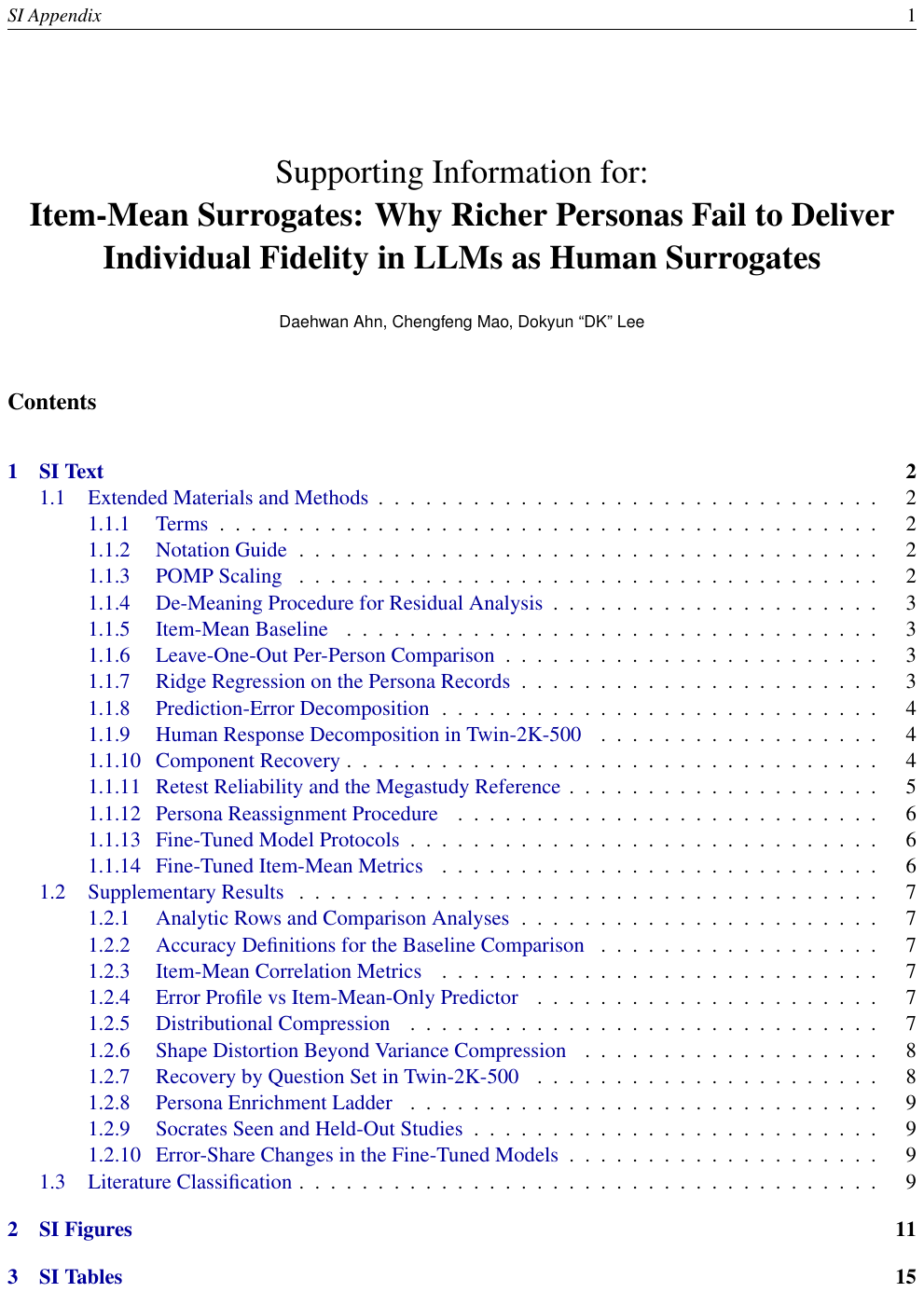}

\end{document}